\documentclass[10pt,twocolumn,letterpaper]{article}

\usepackage{cvpr}
\usepackage{times}
\usepackage{epsfig}
\usepackage{graphicx}
\usepackage{amsmath}
\usepackage{amssymb}
\usepackage{caption}
\usepackage{booktabs}

\usepackage{lipsum}
\usepackage{bbm}
\usepackage{float}
\usepackage{multirow}
\usepackage{enumitem}

\usepackage[pagebackref=true,breaklinks=true,colorlinks,bookmarks=false]{hyperref}

\usepackage{times}
\usepackage{epsfig}
\usepackage{graphicx}
\usepackage{float}
\usepackage{wrapfig}
\usepackage{amsmath,amssymb,amsthm}
\usepackage{algorithm,algorithmicx,algpseudocode}
\usepackage{bm,xspace}
\usepackage{comment}
\usepackage{verbatim}
\usepackage{multirow}
\usepackage{balance}
\usepackage{url}
\usepackage{booktabs}
\usepackage{etoolbox,siunitx}
\usepackage{calc}
\usepackage{pifont,hologo}
\usepackage[usenames, dvipsnames]{color}
\usepackage{nicefrac}

\newcommand{\ra}[1]{\renewcommand{\arraystretch}{#1}}

\usepackage[super]{nth}
\usepackage{nicefrac}
\robustify\bfseries

\DeclareMathSymbol{@}{\mathord}{letters}{"3B}

\def\latex/{\LaTeX}
\def\bibtex/{\hologo{BibTeX}}

\cvprfinalcopy 

\def\cvprPaperID{1058} 

\begin{document}

\title{Single Image Reflection Separation with Perceptual Losses}

\author{Xuaner Zhang \\
UC Berkeley\\
\and
Ren Ng\\
UC Berkeley\\
\and
Qifeng Chen\\
Intel Labs\\
}

\maketitle


\begin{abstract}
We present an approach to separating reflection from a single image. The approach uses a fully convolutional network trained end-to-end with losses that exploit low-level and high-level image information. Our loss function includes two perceptual losses: a feature loss from a visual perception network, and an adversarial loss that encodes characteristics of images in the transmission layers. We also propose a novel exclusion loss that enforces pixel-level layer separation. We create a dataset of real-world images with reflection and corresponding ground-truth transmission layers for quantitative evaluation and model training. We validate our method through comprehensive quantitative experiments and show that our approach outperforms state-of-the-art reflection removal methods in PSNR, SSIM, and perceptual user study. We also extend our method to two other image enhancement tasks to demonstrate the generality of our approach.
\end{abstract}

\begin{figure*}[t!]
\centering
\begin{tabular}{@{}c@{\hspace{3mm}}c@{\hspace{1mm}}c@{\hspace{3mm}}c@{\hspace{1mm}}c@{}}
& Transmission & Reflection & Transmission & Reflection\\
\includegraphics[width=0.19\linewidth]{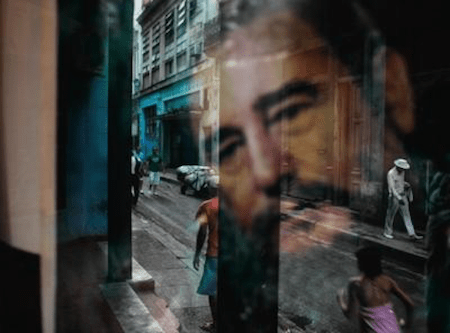}&
\includegraphics[width=0.19\linewidth]{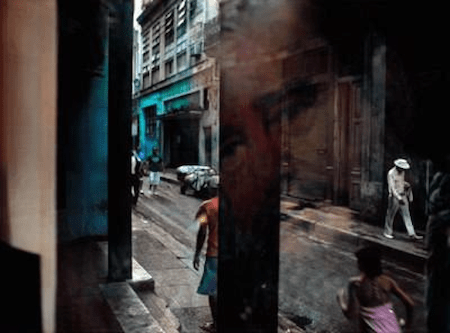}&
\includegraphics[width=0.19\linewidth]{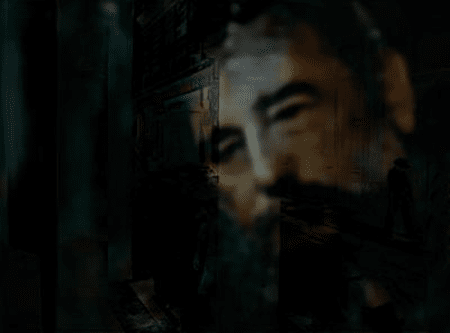}&
\includegraphics[width=0.19\linewidth]{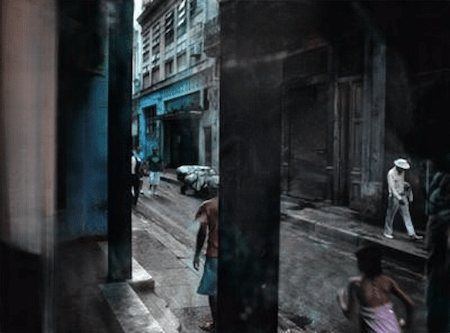}&
\includegraphics[width=0.19\linewidth]{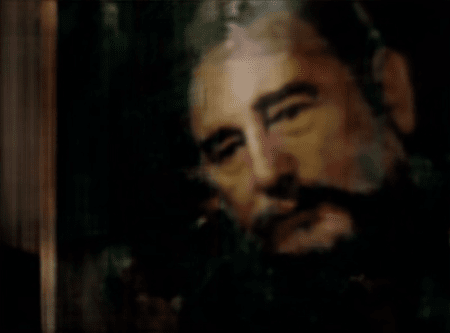}\\
\includegraphics[width=0.19\linewidth]{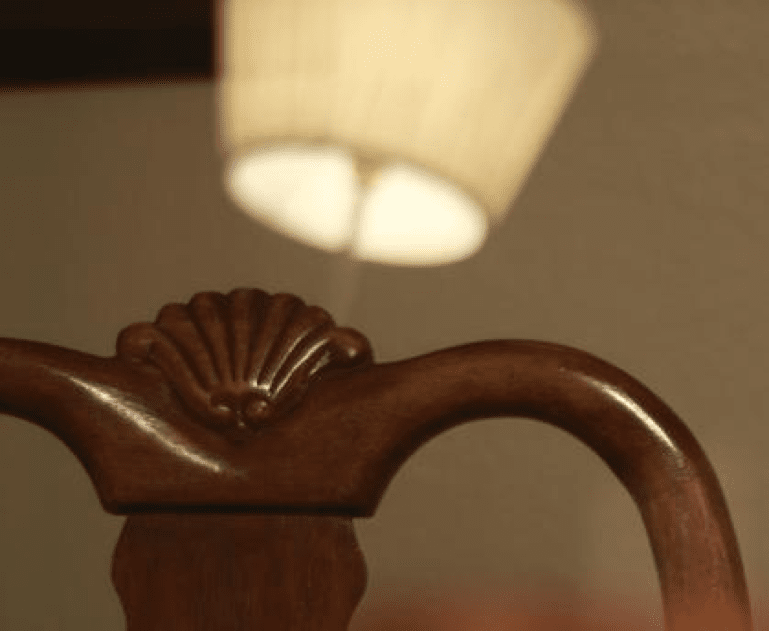}&
\includegraphics[width=0.19\linewidth]{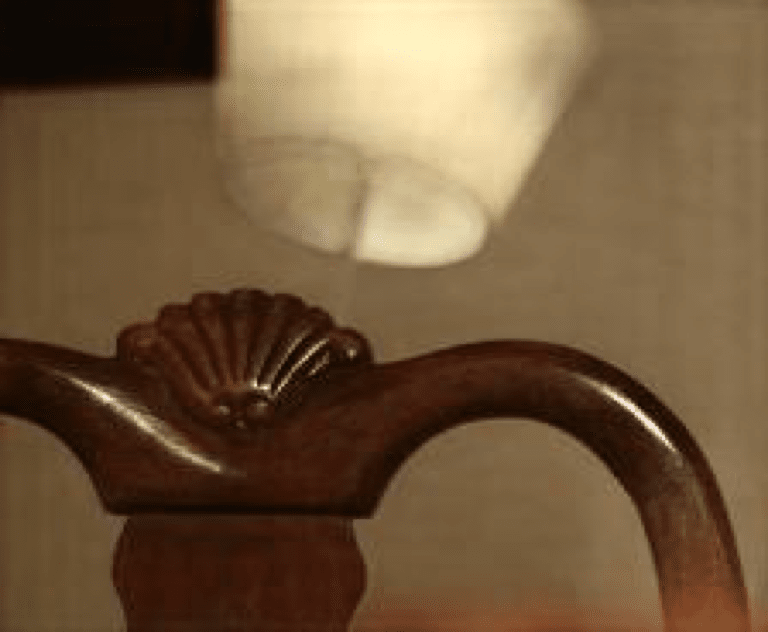}&
\includegraphics[width=0.19\linewidth]{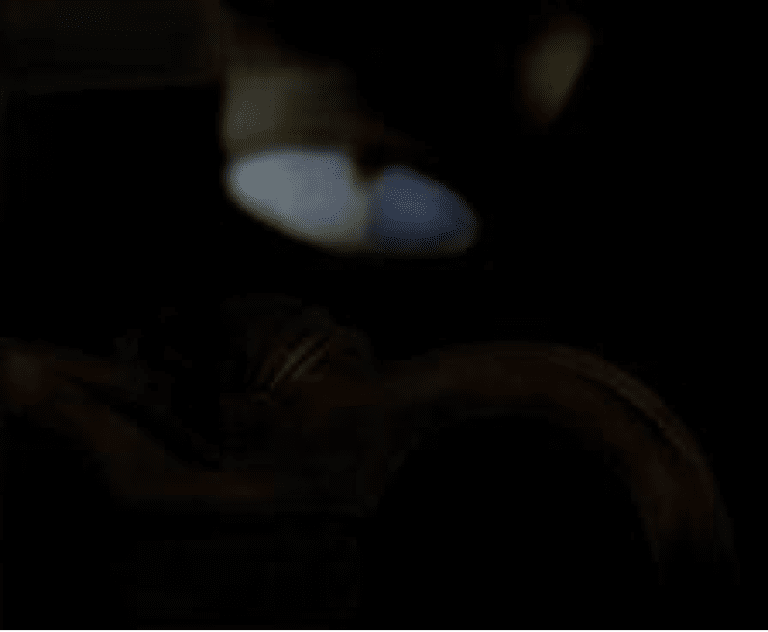}&
\includegraphics[width=0.19\linewidth]{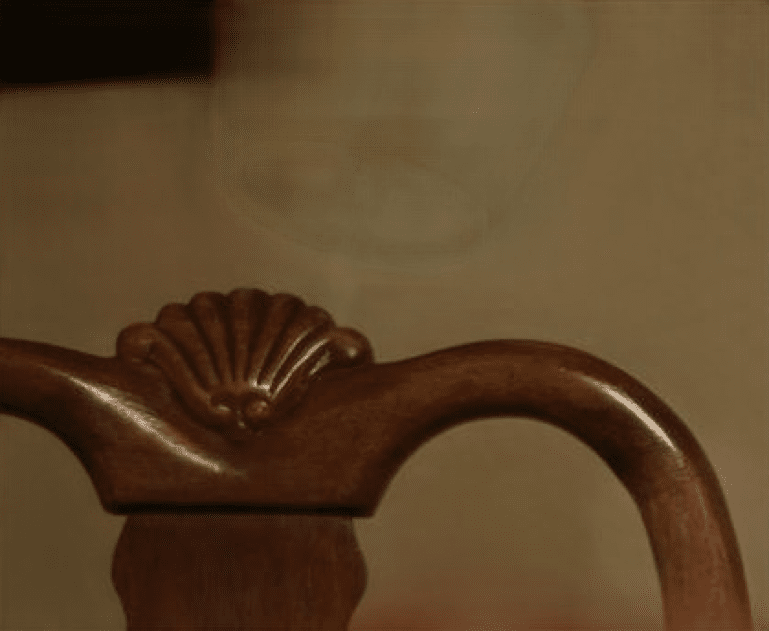}&
\includegraphics[width=0.19\linewidth]{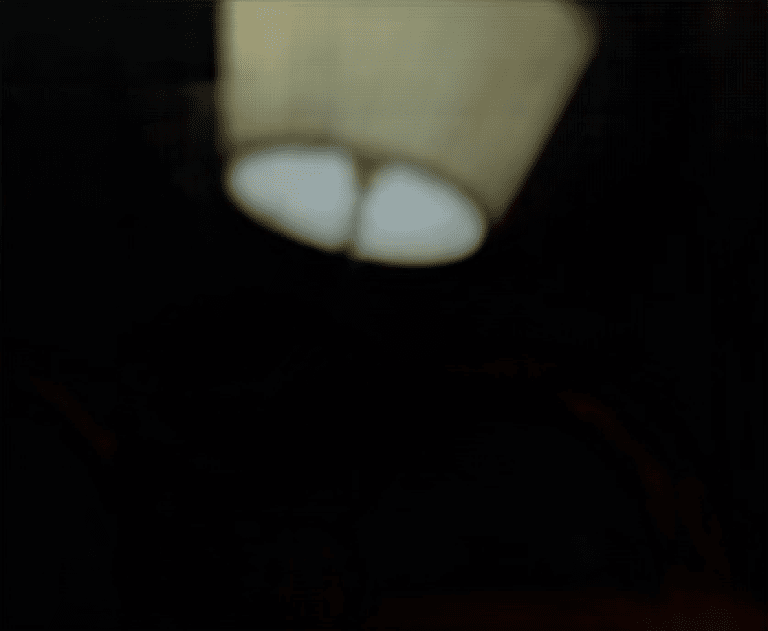}\vspace{1mm}\\
 Input & \multicolumn{2}{c}{CEILNet \cite{Fan2017}} & \multicolumn{2}{c}{Our results} \\
\end{tabular}
\caption{Results by CEILNet \cite{Fan2017} and our approach on real-world images. The top row shows a real image from the CEILNet dataset with a window reflecting a poster of a human face; the bottom row shows an image taken by ourselves, with a lamp as the reflection. From left to right: the input images, CEILNet results and our results. Note that our approach trained to learn both low-level and high-level image statistics successfully removes the reflection layers of the face and lamp, while CEILNet does not.}
\label{fig:teaser}
\vspace{3mm}
\end{figure*}

\section{Introduction}
Reflection from windows and glasses is ubiquitous in the real world,
but it is usually undesirable in photographs. Users often want to extract the hidden clean transmission image by removing reflection from an image. For example, we may have been tempted to take photos through an aquarium glass or skyscraper windows, but reflection can often damage the image quality. Removing reflection from a single image allows us to recover visual content with better perceptibility. Thus, separating the reflection layer and transmission layer from an image --- the \emph{reflection separation problem} --- is an active research area in computer vision.

Let $I \in \mathbb{R}^{m\times n\times 3}$ be the input image with reflection. $I$ can be approximately modeled as the sum of the transmission layer $T$ and the reflection layer $R$: $I=T+R$. Our goal is to recover the transmission layer $T$ given $I$, which is an ill-posed problem without additional constraints or priors.

As the reflection separation problem is ill-posed, prior works often require additional input images and hard-crafted priors. A line of previous research uses multiple images as input or requires explicit user guidance \cite{Guo2014,springer2017reflection,xue2015computational}. Multiple images, however, are not always available in practice, and user guidance is inconvenient and error-prone. Recent researchers proposed methods for reflection removal from a single image \cite{shih2015reflection,li2014single}, but these approaches rely on hand-crafted priors such as ghost cues and relative smoothness which may not generalize to all images with reflection. More recently, CEILNet \cite{Fan2017} uses a deep neural network to train a model with low-level losses on color and edges, but this approach does not directly enable the model to learn high-level semantics which can be highly useful for reflection removal. Low-level information is insufficient for reflection separation when there is color ambiguity or the model needs to "recognize" objects in the image. For example, in Figure \ref{fig:teaser}, our model trained with perceptual losses may have learned the representations of lamps and faces, and thus correctly removes them from the input image, while CEILNet fails to do so.

In this paper, we present a fully convolutional network with perceptual losses that encode both low-level and high-level image information. Our network takes a single image as input and directly synthesizes two images: the reflection layer and the transmission layer. 
We further propose a novel exclusion loss that effectively enforces the separation of transmission and reflection at pixel level. To thoroughly evaluate and train different approaches, we build a dataset that contains real-world images and the ground-truth transmission images. Our dataset covers diverse natural environments including indoor and outdoor scenes. We also use this real-world dataset to compare our approach quantitatively to previous methods. In summary, our main contributions are:
\begin{itemize}
\item We propose to use a deep neural network with perceptual losses for single image reflection separation. We impose perceptual supervision through two losses with different levels of image information: a feature loss from a visual perception network, and an adversarial loss to refine the output transmission layer. 
\item We propose a carefully designed exclusion loss that emphasizes independence of the layers to be separated in the gradient domain.
\item We build a dataset of real-world images for reflection removal with corresponding ground-truth transmission layers. This new dataset enables quantitative evaluation and comparisons among our approach and existing algorithms.
\item Our extensive experiments on real data and synthetic data indicate that our method outperforms state-of-the-art methods in SSIM, PSNR, and a perceptual user study on Amazon Mechanical Turk. Our trained model on reflection separation can be directly applied to two other image enhancement tasks, flare removal and dehazing.
\end{itemize}

\begin{figure*}
\centering
\begin{tabular}{@{}c@{\hspace{0.8mm}}c@{\hspace{0.8mm}}c@{\hspace{0.8mm}}c@{\hspace{0.8mm}}c@{}}
\includegraphics[width=0.195\linewidth]{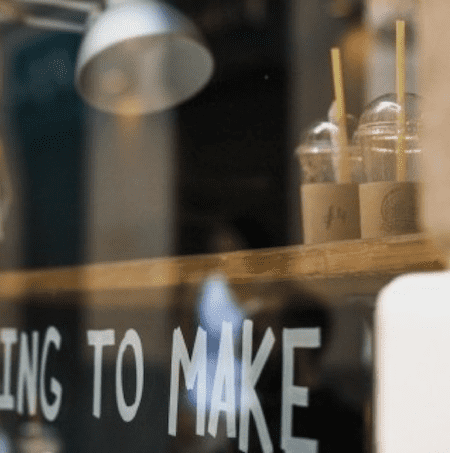}&
\includegraphics[width=0.195\linewidth]{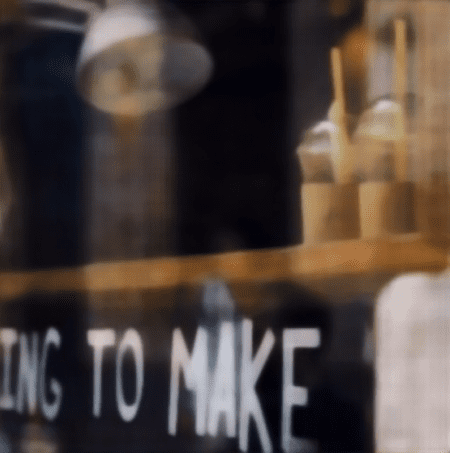}&
\includegraphics[width=0.195\linewidth]{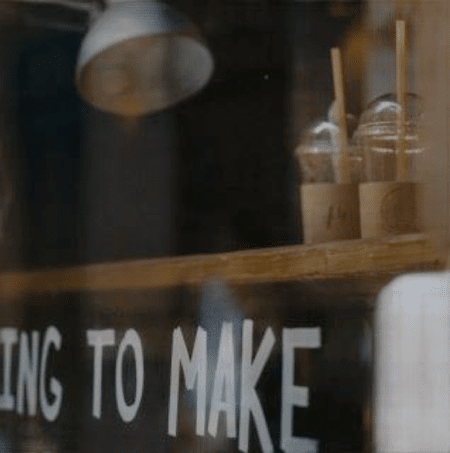}&
\includegraphics[width=0.195\linewidth]{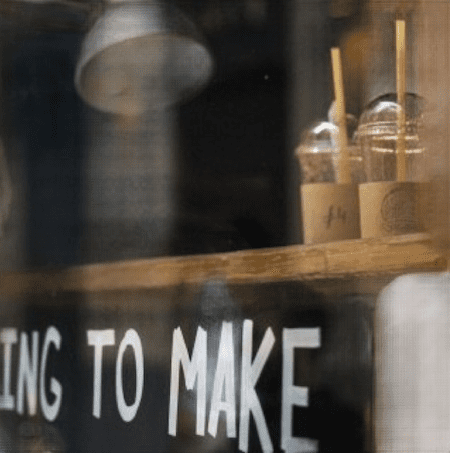}&
\includegraphics[width=0.195\linewidth]{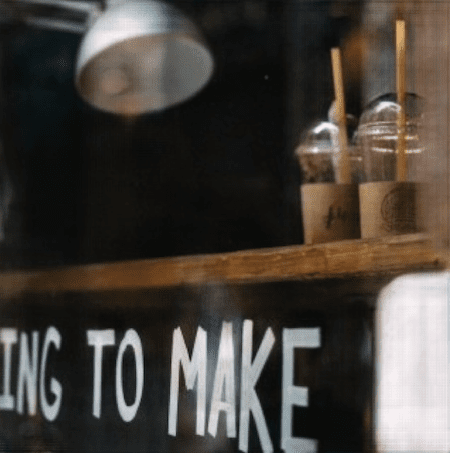}\vspace{1mm}\\
\small (a) Input &\small (b) Without  $L_{\mathrm{feat}}$ & \small (c) Without $L_{\mathrm{adv}}$ & \small (d) Without $L_{\mathrm{excl}}$& \small  (e) Complete model\\
\end{tabular}
\caption{Visual comparisons on the three perceptual loss functions, evaluated on a real-world image. In (b), we replace $L_{\mathrm{feat}}$ with image space $L^1$ loss and observed overly-smooth output. (c) shows artifacts of color degradation and noticeable residuals without $L_{\mathrm{adv}}$. In (d), the lack of $L_{\mathrm{excl}}$ makes the predicted transmission have undesired reflection residuals. Our complete model in (e) is able to produce better and cleaner prediction.}
\label{fig:ablation}
\end{figure*}
\section{Related Work}
\paragraph{Multiple-image methods.}
As the reflection separation problem is ill-posed, most previous work tackles this problem with multiple input images. These multi-image approaches often use motion cues to separate the transmission and reflection layers~\cite{xue2015computational,Guo2014,Li2013,Sun2016,sarel2004separating,Gai2012,szeliski2000layer,han2017reflection}. The motion cues are either inferred from calibrated cameras, or motion parallax that assumes the background and reflection objects have greatly different motion fields. Some other multi-image approaches include the use of flash and no-flash image pairs to improve the flash image with reflection removed~\cite{agrawal2005removing}. Schechner et al.~\cite{Schechner2000} use a sequence of images with different focus settings to separate layers with depth estimation. Kong et al.~\cite{kong2014physically} exploit physical properties of polarization and use multiple polarized images taken with angular filters to find the optimal separation. More recently, Han and Sim~\cite{han2017reflection} tackle the glass reflection removal problem with multiple glass images, assuming that the gradient field in background image is almost constant while the gradient field in reflection varies much more. Although multiple-image methods have shown promising performance in removing reflection, capturing multiple images is sometimes impossible, for example, these methods can not be applied to existing or legacy photographs.
\paragraph{Single-image methods.}
Another line of work considers using a single image with predefined priors. A widely used prior is the natural image gradient sparsity ~\cite{levin2004separating,levin2003learning} to find minimum edges and corners for layer decomposition. The gradient sparsity prior is also explored together with optimal and minimum user assistance to better guide the ill-posed separation problem \cite{Levin2007,springer2017reflection}. A recent work by Arvanitopoulos et al. ~\cite{arvanitopoulossingle} uses the gradient sparsity constraint, combined with a data fidelity term in the Laplacian space to suppress reflection. However, all these approaches rely on low-level heuristics and are limited in cases where a high-level understanding of the image is needed. 

Another prior for reflection separation is that the reflection layer is often out of focus and appears smooth. This is explicitly formulated into an optimization objective by Li and Brown~\cite{li2014single}, in which they penalize large reflection gradients. Although the assumption of relative smoothness is valid, their formulation can break down when the reflection layer has high contrast. Wan et al.~\cite{wan2016depth} propose a variation of this smoothness prior where depth of field is used as guidance for edge labeling and layer separation. Additionally, Shih et al.~\cite{shih2015reflection} focus on a subset of the problem where reflection has ghost effects, and use estimated convolution kernel to optimize for reflection removal.

Fan et al.~\cite{Fan2017} recently propose a deep learning network, the Cascaded Edge and Image Learning Network (CEILNet), for reflection removal. 
They formulate reflection removal as an edge simplification task and learn an intermediate edge map to guide layer separation. 
CEILNet is trained purely with a low-level loss that combines the differences in color space and gradient domain. The main difference between CEILNet and ours is that they did not explicitly utilize perceptual information during training. 

\paragraph{Benchmark datasets.}
A benchmark dataset by Wan et al. \cite{wan2017benchmarking} was proposed recently for reflection removal. The authors collected 1500 real images of 40 scenes in a controlled lab environment by imaging pairs of daily objects and postcards, as well as 100 scenes in natural outdoor environments with three different pieces of glasses. However, the dataset has not been released publicly yet at the time of submission. In order to evaluate among different models quantitatively on real-world images, we collect a dataset of 110 real images with ground truth in natural scene environments.

\section{Overview}
\label{sec:overview}
Given an image $I \in [0,1]^{m \times n \times 3}$ with reflection, our approach decomposes $I$ into a transmission layer $f_T(I;\theta)$ and a reflection layer $f_R(I;\theta)$ using a single network $f(I;\theta)=(f_T(I;\theta),f_R(I;\theta))$, where $\theta$ is the network weights. We train the network $f$ on a dataset $\mathcal{D}=\{(I,T,R)\}$ where $I$ is the input image, $T$ is the transmission layer of $I$, and $R$ is the reflection layer of $I$. 

Our loss function contains three terms: a feature loss $L_{\mathrm{feat}}$ by comparing the images in feature space, and an adversarial loss $L_{\mathrm{adv}}$ for realistic image refinement, an exclusion loss $L_{\mathrm{excl}}$ that enforces separation of the transmission and reflection layers in the gradient domain. Our overall loss function is
\begin{equation}
\label{eq:loss}
L(\theta)=w_1 L_{\mathrm{feat}}(\theta)+w_2 L_{\mathrm{adv}}(\theta)+w_3 L_{\mathrm{excl}}(\theta),
\end{equation}
where we set $w_1=0.1$, $w_2=0.01$ and $w_3=1$ to balance the weight of each term. 

An ideal model for reflection separation should be able to understand contents in an image. 
To train our network $f$ with semantic understanding of the input image, we form hypercolumn features \cite{Hariharan2015} by extracting features from a VGG-19~\cite{simonyan2014very} network pre-trained on the ImageNet dataset \cite{russakovsky2015imagenet}. The benefit of using hypercolumn features is that the input is augmented with useful features that abstract visual perception of a large dataset such as ImageNet. The hypercolumn feature at a given pixel location is a stack of activation units across selected layers of a network at that location. Here, we sampled the layers 'conv1\_2', 'conv2\_2', 'conv3\_2', 'conv4\_2', and 'conv5\_2' in the pre-trained VGG-19 network. The hypercolumn feature has 1472 dimensions in total. We concatenate the input image $I$ with its hypercolumn features as the augmented input for $f$. 

Our network $f$ is a fully convolutional network that has a similar network architecture to the context aggregation network \cite{YuKoltun2016,chen2017fast}. Our network has a large receptive field of $513\times 513$ to effectively aggregate global image information. The first layer of $f$ is a $1\times 1$ convolution to reduce feature dimension (1472+3) to 64. The following 8 layers are $3\times 3$ dilated convolutions. The dilation rate varies from 1 to 128. All the intermediate layers have 64 feature channels. For the last layer we use a linear transformation to synthesize 2 images in the RGB color space. 

We evaluate different methods on the publicly available synthetic and real images from the CEILNet dataset\cite{Fan2017} and the real-world dataset we collected. We compare our method to the state-of-the-art reflection removal approach CEILNet~\cite{Fan2017}, an optimization based approach~\cite{li2014single}, and Pix2pix \cite{isola2017image}, a general framework for image translation.
\section{Training}
\label{sec:method}

\subsection{Feature loss}
We use a feature loss to measure the difference between our predicted transmission layer and the ground-truth transmission in feature space. As the aforementioned observation in Figure \ref{fig:teaser} shows, semantic reasoning about the scene would benefit the task of reflection removal. A feature loss that combines low-level and high-level features  from a perception network would serve our purpose. Feature loss has also been successfully applied to other tasks such as image synthesis and style transfer \cite{chen2017photographic,Gatys2016,Ledig2016,Johnson2016}.

Here, we compute the feature loss by feeding the predicted image layer and the ground truth through a pre-trained VGG-19 network $\Phi$. We compute the $L^1$ difference between $\Phi(f_T(I;\theta)$ and $\Phi(T)$ in selected feature layers: \begin{equation}
L_{\mathrm{feat}}(\theta)=\sum_{(I,T)\in \mathcal{D}}{\sum_l{\lambda_l\|\Phi_l(T)-\Phi_l(f_T(I;\theta))\|_1}},
\end{equation}
where $\Phi_l$ indicates the layer $l$ in the VGG-19 network. The weights $\{\lambda_l\}$ are used to balance different terms in the loss function. We select the layers 'conv1\_2', 'conv2\_2', 'conv3\_2', 'conv4\_2', and 'conv5\_2' in the VGG-19 network.


\subsection{Adversarial loss} \label{par:gan} 
During the course of our research, we find that transmission image can suffer from unrealistic color degradation and undesirable subtle residuals without an adversarial loss. We adopted the conditional GAN~\cite{isola2017image} for our model. Our generator would be $f_T(I;\theta)$. The architecture of our discriminator, denoted as $D$, has 4 layers and 64 feature channels wide. The discriminator tries to discriminate between patches in the real transmission images and patches given by $f_T(I;\theta)$ conditioned on $I$. The goal is to let the network $D$ learn a suitable loss function for further refining layer separation, and to push the predicted transmission layers toward the domain of real reflection-free images. 

Loss for the discriminator $D$ is:
\begin{equation}
\sum_{(I,T)\in \mathcal{D}}{\log{D(I, f_T(I;\theta))}-\log{D(I, T)}},
\end{equation}
where $D(I, x)$ outputs the probability that $x$ is a natural transmission image given the input image $I$. Then our adversarial loss is:
\begin{equation}
L_{\mathrm{adv}}(\theta)=\sum_{I \in \mathcal{D}}{-\log D(I,f_T(I;\theta))}.
\end{equation}
We optimize over $-\log{D(I, f_T(I;\theta))}$ instead of $\log{(1-D(I, f_T(I;\theta)))}$ for better gradient performance~\cite{goodfellow2014generative}.

\begin{figure}[t]
\centering
\begin{tabular}{@{}c@{\hspace{0.8mm}}c@{\hspace{0.8mm}}c@{\hspace{0.8mm}}c@{}}
\includegraphics[width=0.24\linewidth]{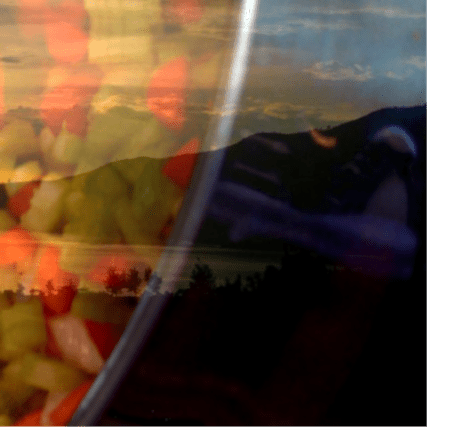}&
\includegraphics[width=0.24\linewidth]{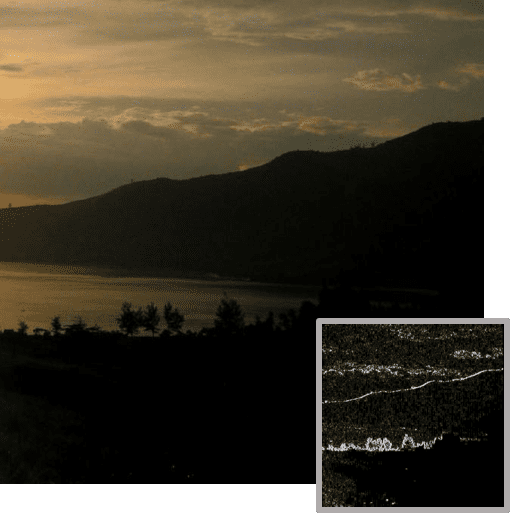}&
\includegraphics[width=0.24\linewidth]{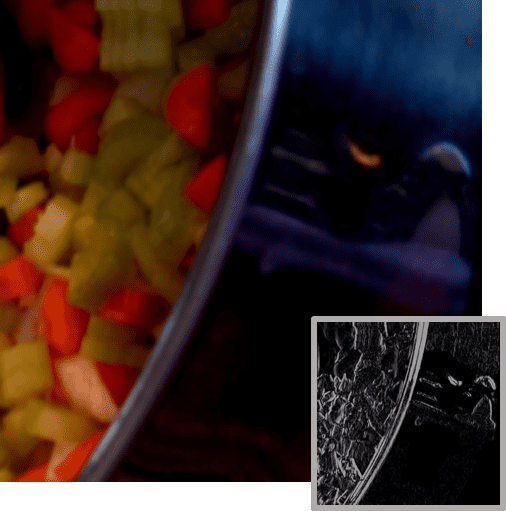}&
\includegraphics[width=0.24\linewidth]{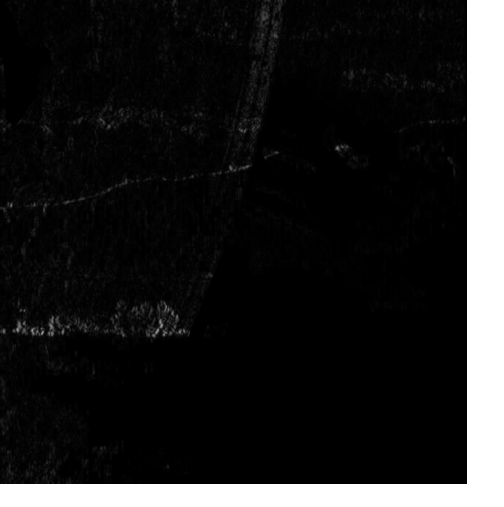}\\
\small I & \small T & \small R & \small $\Psi(T,R)$\\
\includegraphics[width=0.24\linewidth]{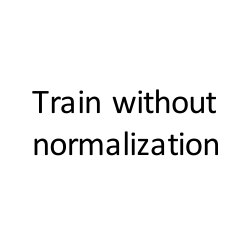}&
\includegraphics[width=0.24\linewidth]{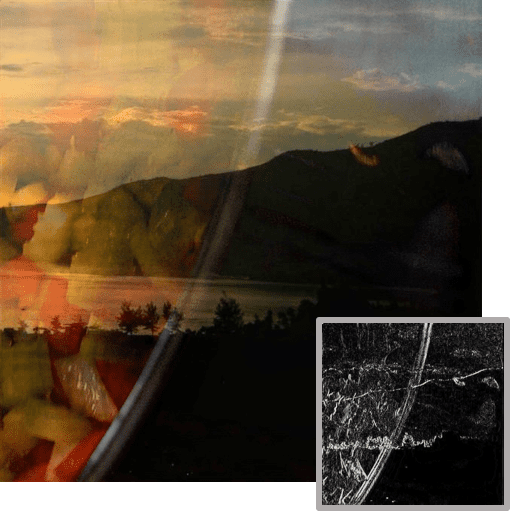}&
\includegraphics[width=0.24\linewidth]{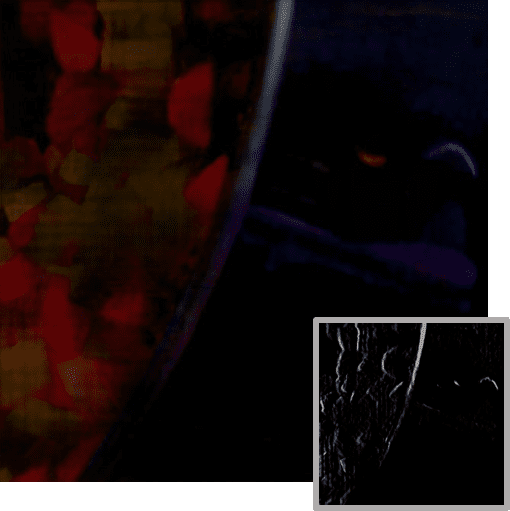}&
\includegraphics[width=0.24\linewidth]{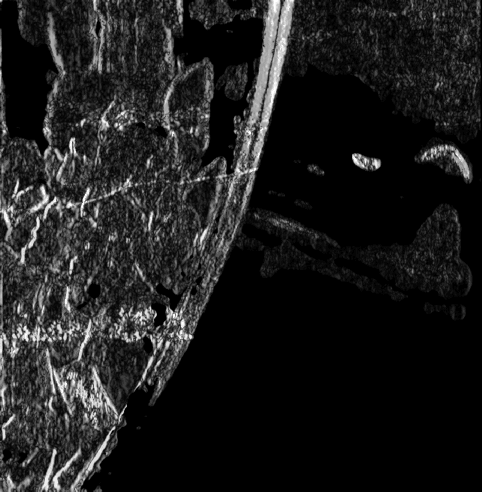}\\
\includegraphics[width=0.24\linewidth]{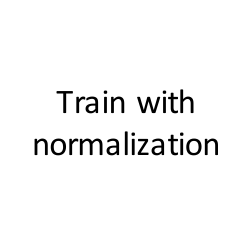}&
\includegraphics[width=0.24\linewidth]{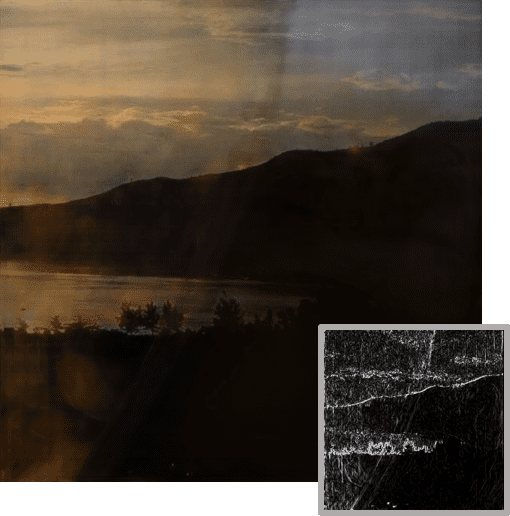}&
\includegraphics[width=0.24\linewidth]{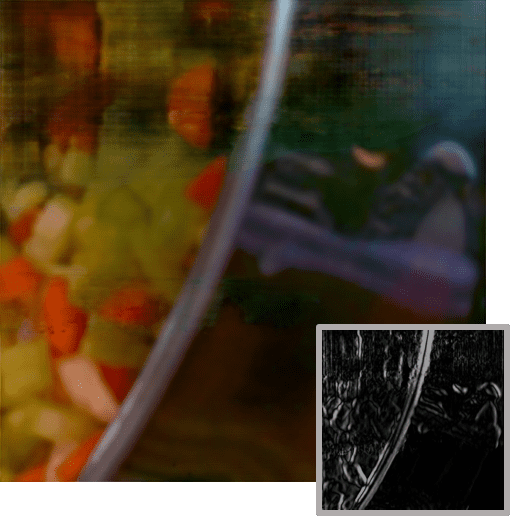}&
\includegraphics[width=0.24\linewidth]{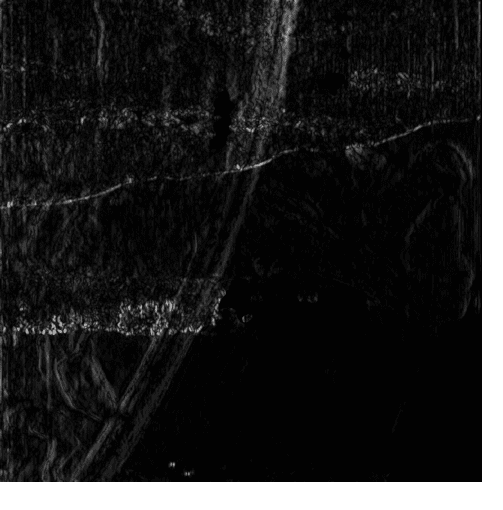}\\
& \small $f_T$& \small $f_R$& \small  $\Psi(f_T,f_R)$\\
\end{tabular}
\caption{\normalsize Visual comparisons of training with and without gradient normalization. In the middle two columns, the small window at the right bottom corner of each image shows the gradient magnitude of each image. In the rightmost column, $\Psi$ denotes the normalized gradient product formulated in Equation \ref{eq:loss_grad}. The first row left to right shows: input, ground truth transmission $T$, ground truth reflection $R$, and $\Psi$. $\Psi(T,R)$ is close to zeros indicating that the gradient fields of $T$ and $R$ are not correlated. The middle row shows results trained with no normalization in the gradient fields. We observe that the reflection prediction trained without normalization is heavily suppressed. Bottom row shows results trained with gradient normalization with better reflection separation.}
\label{fig:norm}
\end{figure}
\subsection{Exclusion loss}
We further propose an exclusion loss in the gradient domain to better separate the reflection and transmission layers. 
We explore the relationship between the two layers through analysis of the edges in the two layers. Our key observation is that the edges of the transmission and the reflection layers are unlikely to overlap. An edge in $I$ should be caused by either $T$ or $R$, but not both. Thus we minimize the correlation between the predicted transmission and reflection layers in the gradient domain. We formulate the exclusion loss as the product of normalized gradient fields of the two layers at multiple spatial resolutions :
\begin{eqnarray}
L_{\mathrm{excl}}(\theta)&=&\sum_{I \in \mathcal{D}}\sum_{n=1}^{N}\|\Psi(f_T^{\downarrow n}(I;\theta),f_R^{\downarrow n}(I;\theta))\|_F,\\
\Psi(T,R)&=&\tanh(\lambda_T|\nabla T|)\odot \tanh(\lambda_R |\nabla R|),
\label{eq:loss_grad}
\end{eqnarray}
where $\lambda_T$ and $\lambda_R$ are normalization factors, $\|\cdot\|_F$ is the Frobenius norm, $\odot$ denotes element-wise multiplication, and $n$ is the image downsampling factor: the images $f_T$ and $f_R$ are downsampled by a factor of $2^{n-1}$ with bilinear interpolation. We set $N=3$, $\lambda_T=\sqrt{\frac{\|\nabla{R}\|_F}{\|\nabla{T}\|_F}}$, and $\lambda_R=\sqrt{\frac{\|\nabla{T}\|_F}{\|\nabla{R}\|_F}}$ in our experiments.

Note that the normalization factors $\lambda_T$ and $\lambda_R$ are critical in Equation \ref{eq:loss_grad}, since the transmission and reflection layers may contain unbalanced gradient magnitudes. The reflection layer can be either blurred with low intensity and thus consists of small gradients, or it could reflect very bright light and composes brightest spots in the image, which produces high contrast reflection and thus large gradients. 
A scale discrepancy between $|\nabla{T}|$ and $|\nabla{R}|$ would cause unbalanced updates to the two layer predictions. We observe that without proper normalization factors, the network would suppress the layer with a smaller gradient update rate to close to zero. A visual comparison of results with and without normalization is shown in Figure ~\ref{fig:norm}.

$L_{\mathrm{excl}}$ is effective in separating the transmission and reflection layers at the pixel level. If we disable $L_{\mathrm{excl}}$ in our model, some residual reflection may remain visible in the output transmission image, as shown in Figure \ref{fig:ablation} (d).


\subsection{Implementation}
Given the ground-truth reflection layer $R$, we can further constrain $f_R(I;\theta)$ with $R$. Reflection layer is usually not in focus and thus blurry. We simply add a $L^1$ loss in color space to constrain $f_R(I;\theta)$:
\begin{equation}
L_R(\theta)=\sum_{(I,R)\in \mathcal{D}}\|f_R(I;\theta)-R\|_1.
\end{equation}
We train the network $f$ by minimizing $(L+L_\mathrm{R})$ on synthetic and real data jointly. Note that we disable $L_\mathrm{R}$ when training on a real-world image as it is difficult to estimate $R$ precisely. We tried computing $R=I-T$ but $R$ sometimes contains significant artifacts because $I=R+T$ may not hold when $I$ is overexposed.

For the training data, we use 5000 synthetic images and extract 500 image patches from 90 real-world training images with random resolutions between 256p and 480p. To further augment the data, we randomly resize image patches while keeping the original aspect ratio. We train for 250 epochs with batch size 1 on an Nvidia Titan X GPU and weights are updated using the Adam optimizer \cite{KingmaBa2015} with a fixed learning rate of $10^{-4}$. 
\begin{figure}[t!]
\vspace{6mm}
\centering
\includegraphics[width=1.05\linewidth]{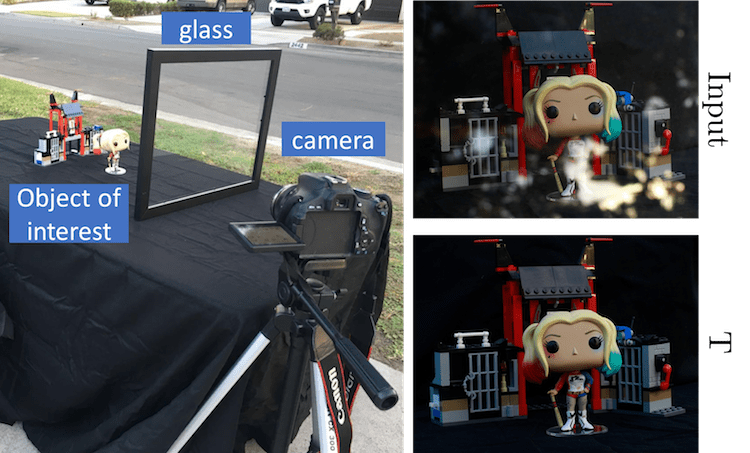}
\vspace{0.1mm}
\caption{Real data collection setup and captured images. We capture two images with and without the glass with same camera settings in a static scene. Right column from top to bottom: captured image with reflection and the ground-truth transmission image $T$.}
\label{fig:setup}
\end{figure}
\section{Dataset}
\label{sec:data}
\subsection{Synthetic data}
To create synthetic images with reflection, we choose 5000 random pairs of images from Flickr: one outdoor image and one indoor image for each pair. We use an image (either indoor or outdoor) as the transmission layer and the other image as the reflection layer. We assume the transmission and reflection layers locate on different focal planes so that the two layers exhibit noticeable different blurriness. This is a valid assumption in real-life photography, where the object of interest (e.g. artwork through museum windows) is often in the transmission layer and is set to be in focus. In addition, reflection could be intentionally blurred by shooting with a wide aperture. We use this assumption to create a synthetic dataset, by applying a Gaussian smoothing kernel with a random kernel size in the range of 3 to 17 pixels to the reflection image.

Our image composition approach is similar to the one proposed by Fan et al. \cite{Fan2017}, but our forward model has the following differences. We remove gamma correction from the images and operate in linear space to better approximate the physical formation of images. Instead of fixing the intensity decay on $R$, we apply variation to the intensity decay since we observe that reflection in real images could have comparable or higher intensity level than the transmission layer. We apply slight vignette centered at random position in the reflection layer, which simulates the scenario when camera views the reflection from oblique angles.

\begin{table}[ht]
\centering
\setlength{\tabcolsep}{3mm}
\ra{1.15}
\begin{tabular}{@{}l@{\hspace{7mm}}c@{\hspace{4mm}}c@{\hspace{7mm}}c@{\hspace{4mm}}c@{}}
\toprule
& \multicolumn{2}{c@{\hspace{12mm}}}{Synthetic} & \multicolumn{2}{c}{Real}\\
Method & SSIM & PSNR & SSIM & PSNR \\
\midrule
Input & 0.689 & 15.09 & 0.697 & 17.66 \\
Pix2pix~\cite{isola2017image} & 0.583 & 14.47 & 0.648 & 16.92\\
Li and Brown~\cite{li2014single} & 0.742 & 15.30 & 0.750 & 18.29\\
CEILNet~\cite{Fan2017} & 0.826 & 20.47 & 0.762 & 19.04\\
Ours & \textbf{0.853} & \textbf{22.63} & \textbf{0.821} & \textbf{21.30}\\
\bottomrule
\end{tabular}
\caption{Quantitative comparison results among our method and 3 other previous methods. We evaluated on synthetic data provided by CEILNet~\cite{Fan2017}, and our real image test set. We also provide a trivial baseline that takes the input image as the result transmission image.}
\label{table:quant_other}
\end{table}

\subsection{Real data}
\label{sec:real_data}
At the time of developing this work, there is no publicly available benchmark with ground-truth transmission to evaluate different reflection removal approaches on real data.  
We collected a dataset of 110 real image pairs: image with reflection and its corresponding ground-truth transmission image. The images with reflection were taken with a Canon 600D camera on a tripod with a portable glass in front of the camera. The ground-truth transmission layer was captured when the portable glass was removed. Each image pair was taken with the same exposure setting. Our setup for data capture is shown in Figure \ref{fig:setup}. We captured the dataset with the following considerations:
\begin{itemize}[noitemsep,topsep=0pt]
\item environments: indoor and outdoor;
\item lighting conditions: skylight, sunlight, and incandescent;
\item camera viewing angles: front view and oblique view;
\item and camera apertures (affecting the reflection blurriness): $f$/2.0 --- $f$/16.
\end{itemize}
We split the dataset randomly into a training set and a test set. We extract 500 patches from 90 training images for training and use 20 images for quantitative evaluation.


\section{Experiments}
\label{sec:exp}

\subsection{Comparison to prior work}
We compare our model to CEILNet \cite{Fan2017}, the layer separation method by Li and Brown \cite{li2014single}, and Pix2pix \cite{isola2017image}. We evaluated different methods on the publicly available synthetic images from the CEILNet dataset \cite{Fan2017} and the real images from the test set of our real-world dataset. 

Our model is only trained on our generated synthetic dataset and the training set of our real-world dataset. For CEILNet, we evaluate its pre-trained model on the CEILNet synthetic images. To evaluate CEILNet on our real data, we fine-tune its model with our real training images (otherwise it performs poorly). We evaluate the approach of Li and Brown \cite{li2014single} with the provided default parameters. Pix2pix is a general image translation model, we train its model on our generated synthetic dataset and the training set of our collected real dataset.

\begin{table}
\centering
\setlength{\tabcolsep}{3mm}
\ra{1.15}
\begin{tabular}{@{}lc@{}}
\toprule
& Preference rate \\
\midrule
Ours$>$CEILNet \cite{Fan2017}& 84.2\%\\
Ours$>$Li and Brown \cite{li2014single} & 87.8\%\\
\bottomrule
\end{tabular}
\caption{User study results. 
The preference rate shows the percentage of comparisons in which users prefer our results.}
\label{tab:user_study}
\end{table}

The quantitative results are shown in Table \ref{table:quant_other}. We compute the PSNR and SSIM between the result transmission images of different methods and ground-truth transmission layer. We demonstrate strong quantitative performance over previous works on both synthetic and real data. 

We also conduct a user study on Amazon Mechanical Turk, following the protocol by Chen and Koltun \cite{chen2017photographic}. During the user study, each user is presented with a input real-world image with reflection, our predicted transmission image, and the predicted transmission image by a baseline in the same row. Then the user needs to choose an output image that is closer to the reflection-free version of the input image between the two predicted transmission images. There are 80 real-world images for comparisons from our dataset and the CEILNet dataset. The results are reported in Table \ref{tab:user_study}. $84.2\%$ of the comparisons to CEILNet and $87.8\%$ of the comparisons to Li and Brown have our results rated to contain less reflection. The results are statistically significant with $p < 10^{-3}$ and 20 users participate in the user study.

More experimental details and results are reported in the supplement.

\begin{figure*}[h!]
\begin{tabular}{@{}c@{\hspace{0.8mm}}c@{\hspace{0.8mm}}c@{\hspace{0.8mm}}c@{\hspace{0.8mm}}c@{\hspace{0.8mm}}c@{}}
& & Transmission & Reflection & Transmission & Reflection\\
\includegraphics[width=0.163\linewidth]{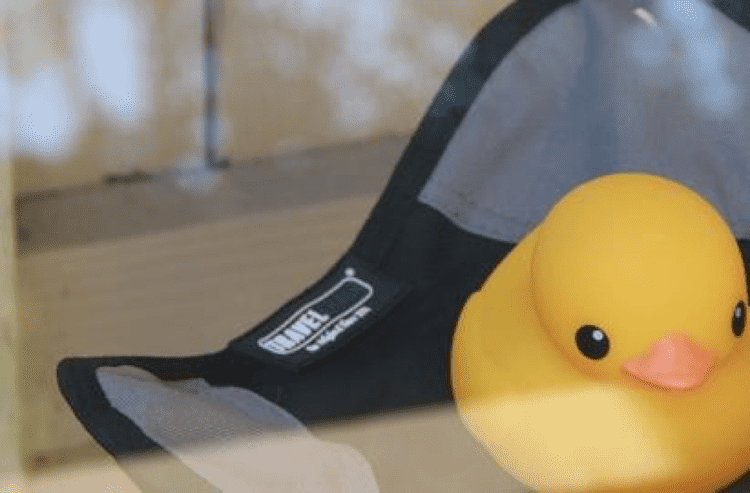}&
\includegraphics[width=0.163\linewidth]{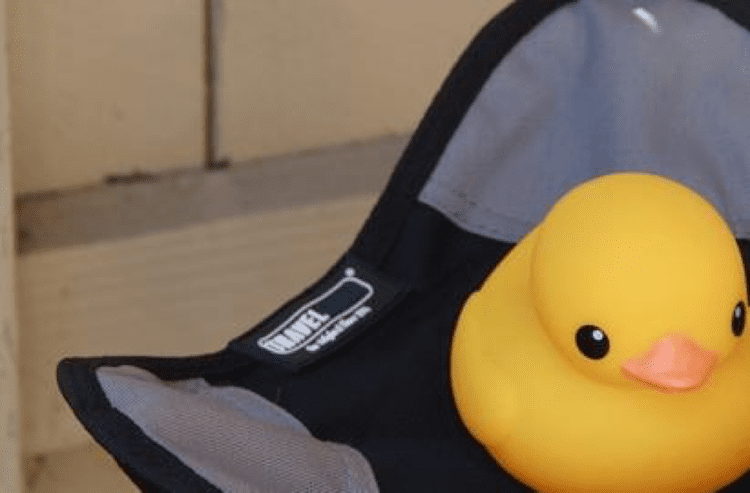}&
\includegraphics[width=0.163\linewidth]{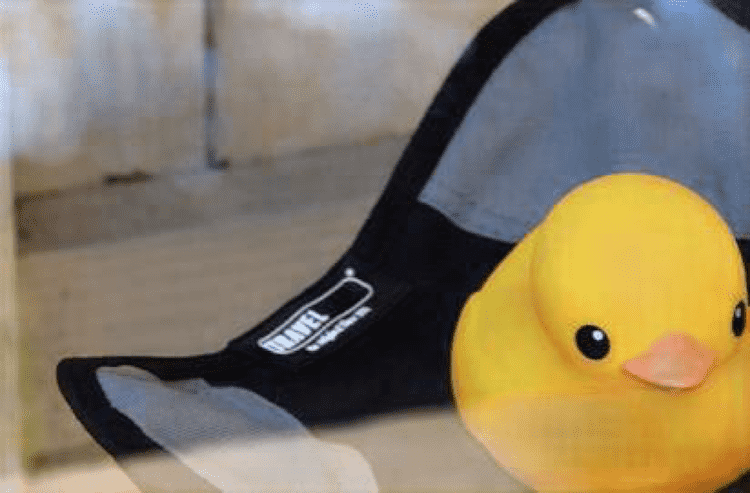}&
\includegraphics[width=0.163\linewidth]{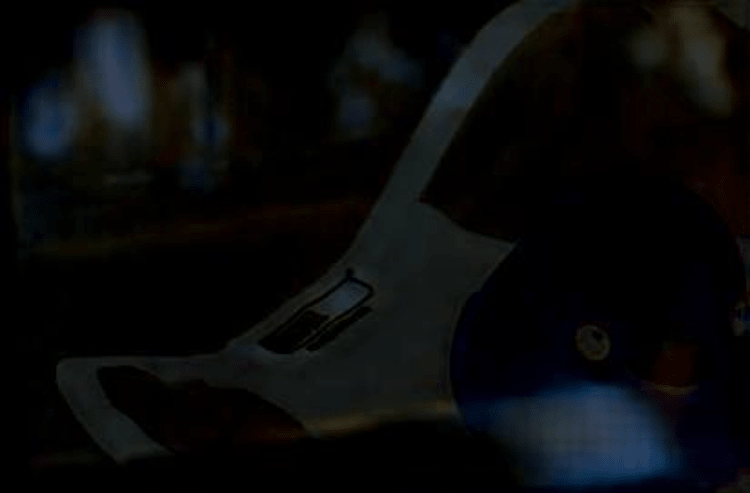}&
\includegraphics[width=0.163\linewidth]{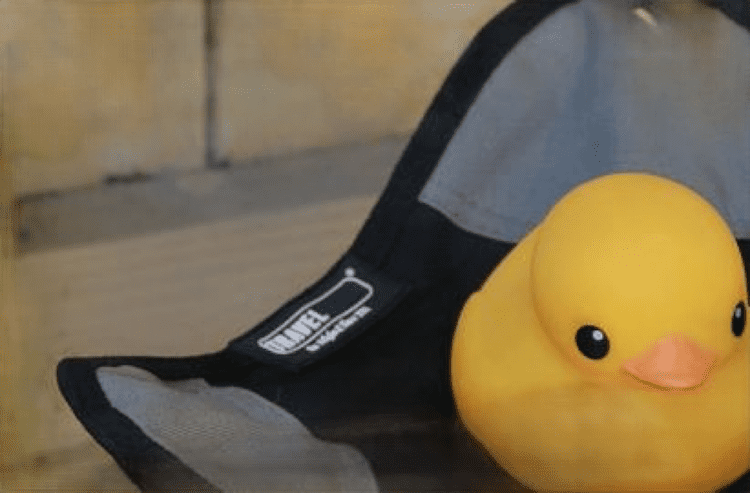}&
\includegraphics[width=0.163\linewidth]{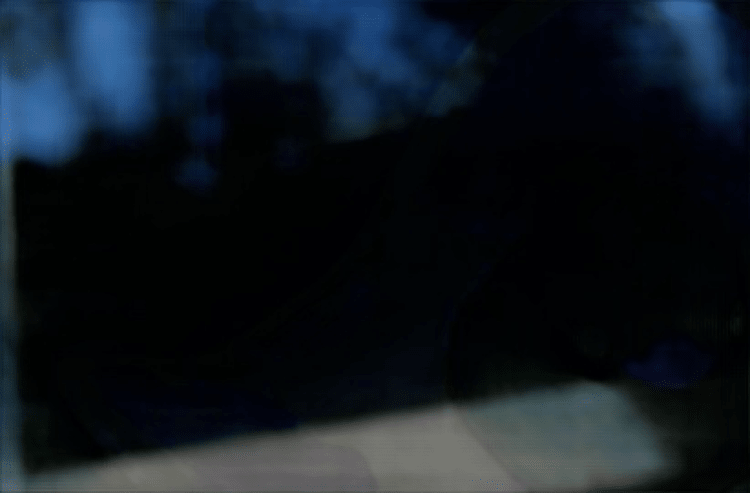}
\vspace{1mm}\\
\includegraphics[width=0.163\linewidth]{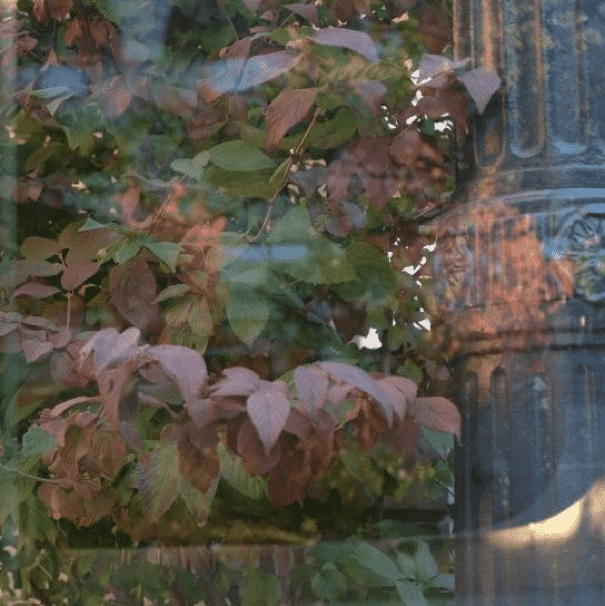}&
\includegraphics[width=0.163\linewidth]{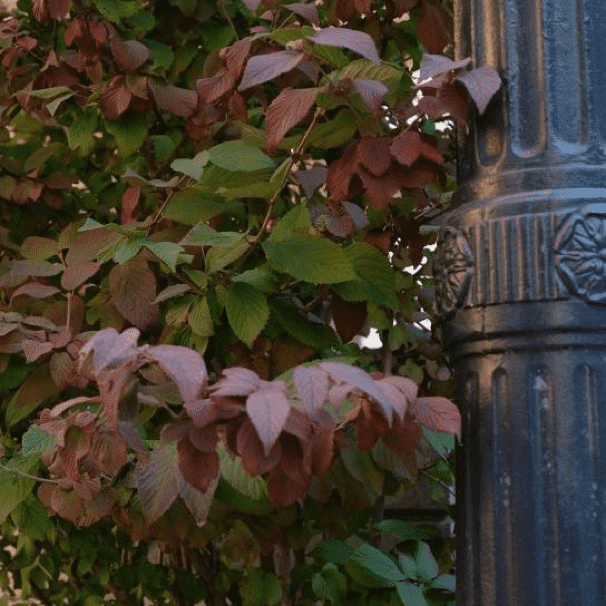}&
\includegraphics[width=0.163\linewidth]{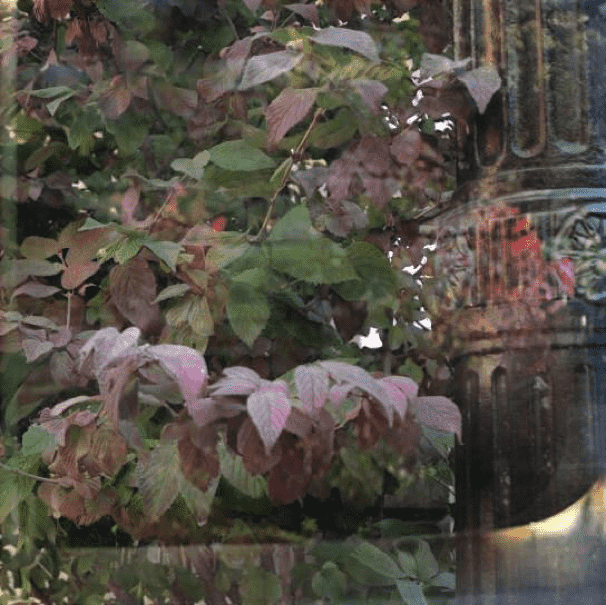}&
\includegraphics[width=0.163\linewidth]{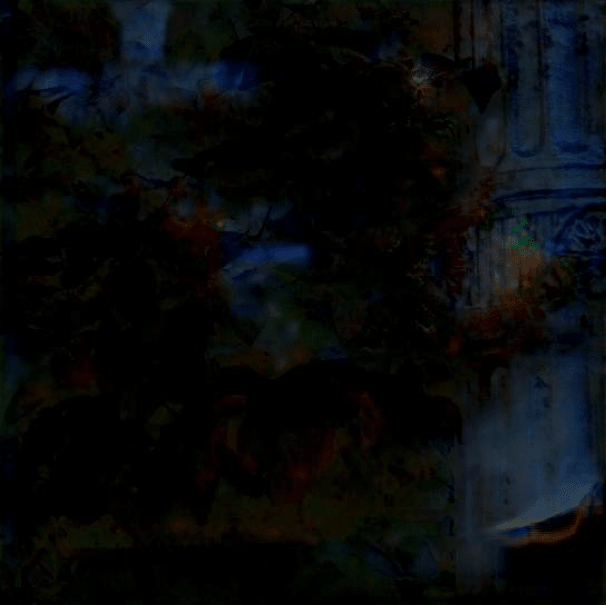}&
\includegraphics[width=0.163\linewidth]{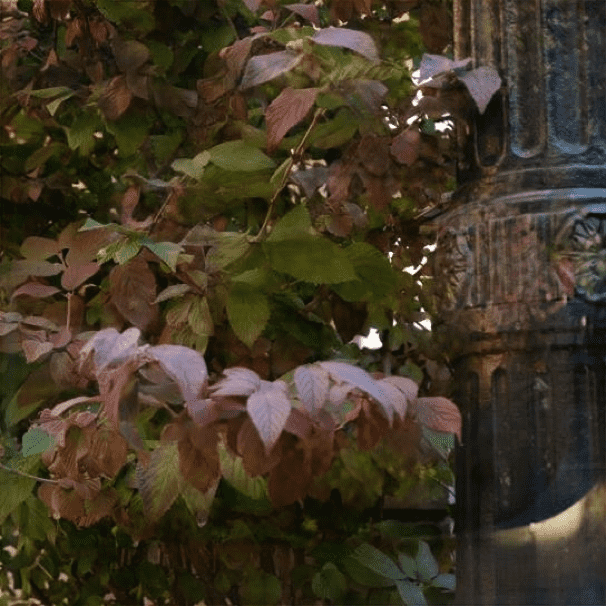}&
\includegraphics[width=0.163\linewidth]{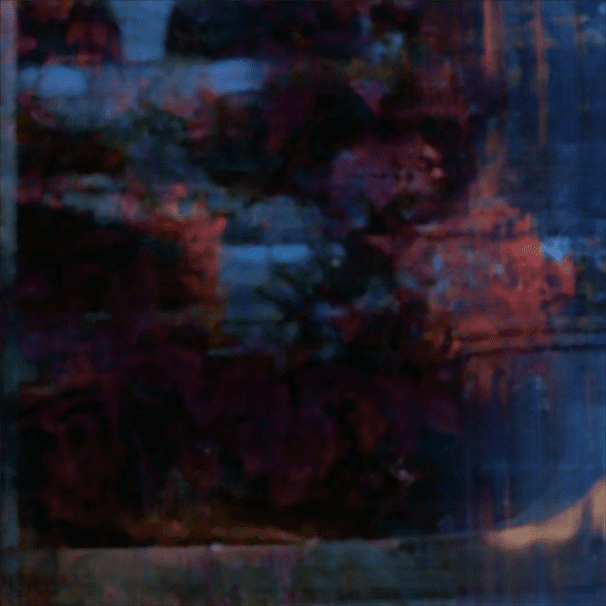}
\vspace{1mm}\\
\includegraphics[width=0.163\linewidth]{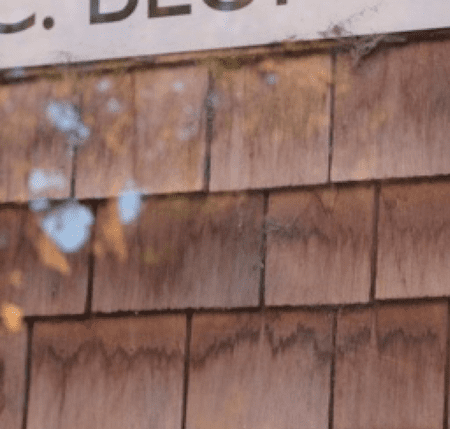}&
\includegraphics[width=0.163\linewidth]{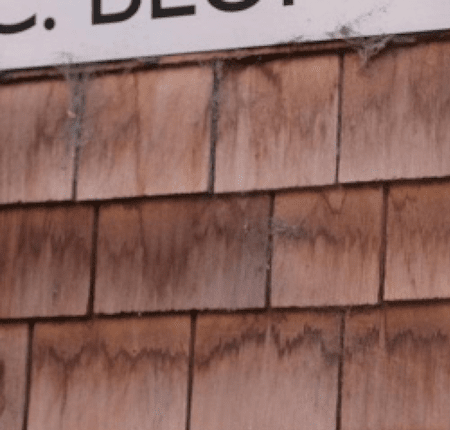}&
\includegraphics[width=0.163\linewidth]{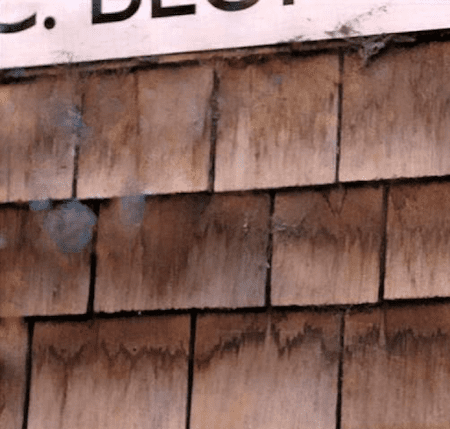}&
\includegraphics[width=0.163\linewidth]{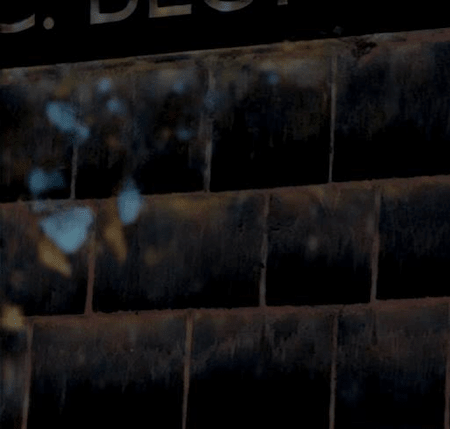}&
\includegraphics[width=0.163\linewidth]{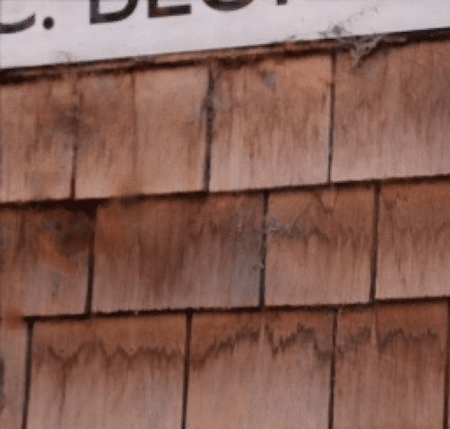}&
\includegraphics[width=0.163\linewidth]{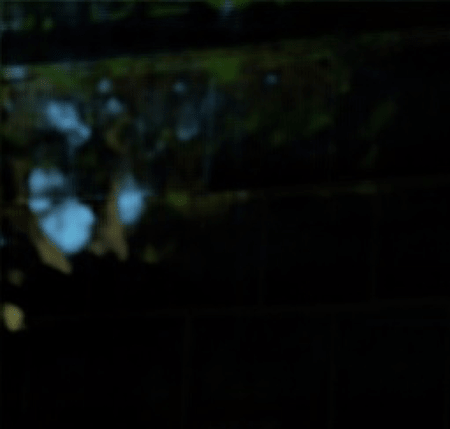}
\vspace{1mm}\\
\small Input & \small Ground-truth T  & \multicolumn{2}{c}{CEILNet~\cite{Fan2017} } & \multicolumn{2}{c}{Our results}
\end{tabular}
\caption{Visual results comparison between CEILNet~\cite{Fan2017} and our method, evaluated on real images from our dataset described in Section~\ref{sec:real_data}. From left to right: input, ground truth transmission layer, CEILNet~\cite{Fan2017} predictions and our predictions. Notice that our method produces better and cleaner predictions in both the transmission and reflection layers. Additional results are provided in the supplement.}
\label{fig:qual_real1}
\vspace{3mm}
\centering
\begin{tabular}{@{}c@{\hspace{0.8mm}}c@{\hspace{0.8mm}}c@{\hspace{0.8mm}}c@{\hspace{0.8mm}}c@{\hspace{0.8mm}}c@{\hspace{0.8mm}}c@{}}
& Transmission & Reflection & Transmission & Reflection & Transmission & Reflection\\
\includegraphics[width=0.14\linewidth]{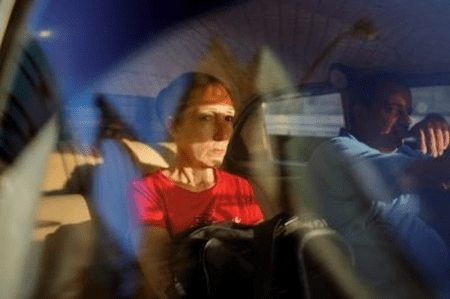}&
\includegraphics[width=0.14\linewidth]{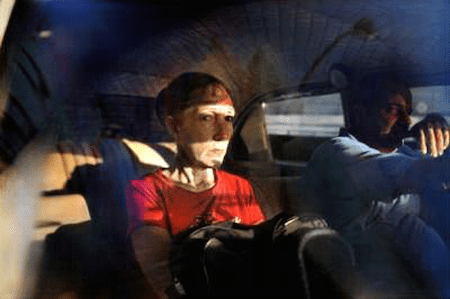}&
\includegraphics[width=0.14\linewidth]{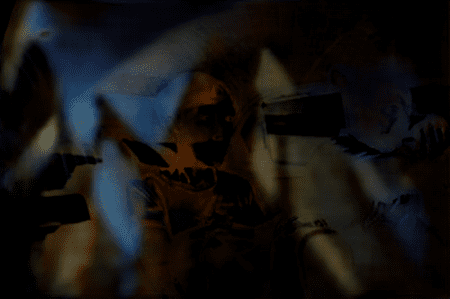}&
\includegraphics[width=0.14\linewidth]{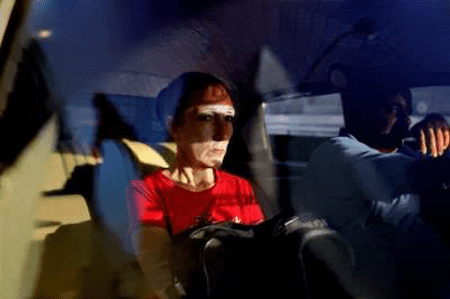}&
\includegraphics[width=0.14\linewidth]{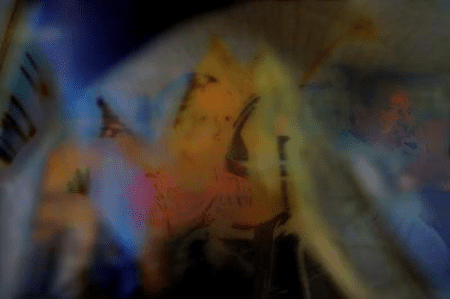}&
\includegraphics[width=0.14\linewidth]{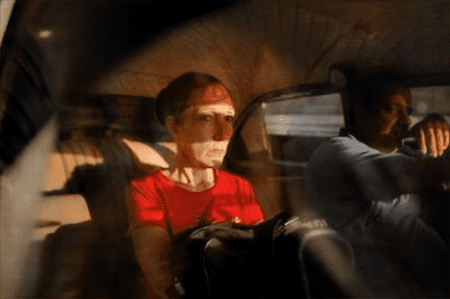}&
\includegraphics[width=0.14\linewidth]{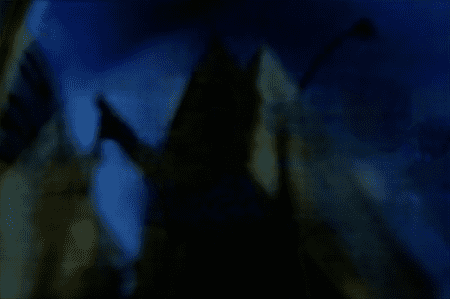}
\vspace{1mm}\\
\includegraphics[width=0.14\linewidth]{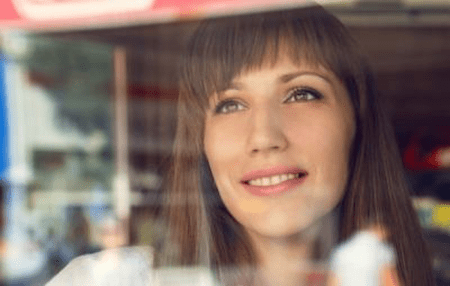}&
\includegraphics[width=0.14\linewidth]{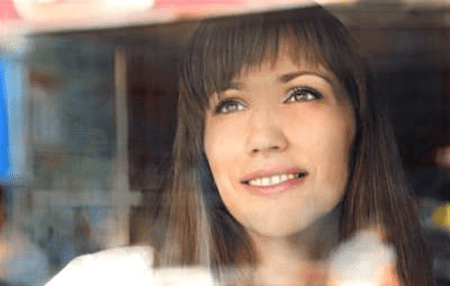}&
\includegraphics[width=0.14\linewidth]{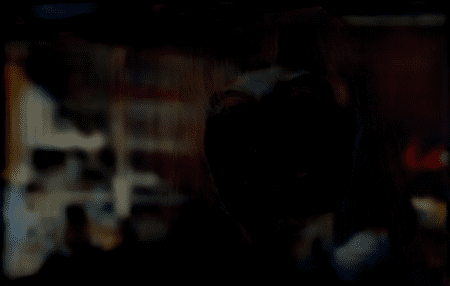}&
\includegraphics[width=0.14\linewidth]{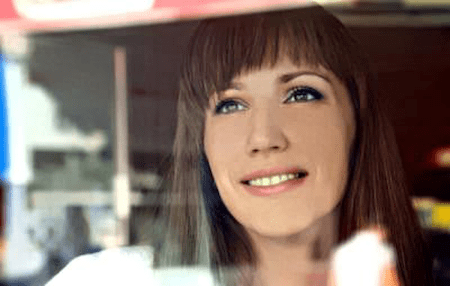}&
\includegraphics[width=0.14\linewidth]{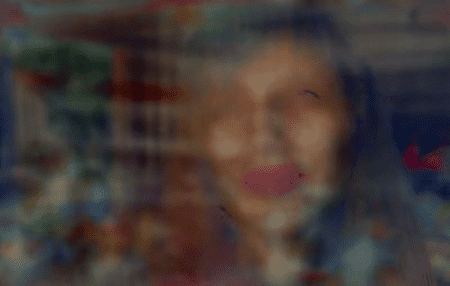}&
\includegraphics[width=0.14\linewidth]{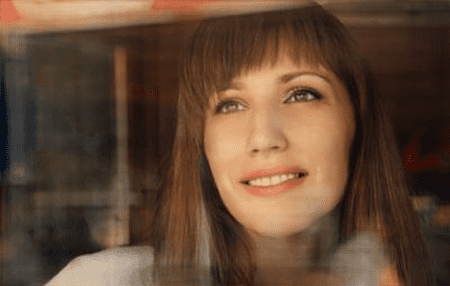}&
\includegraphics[width=0.14\linewidth]{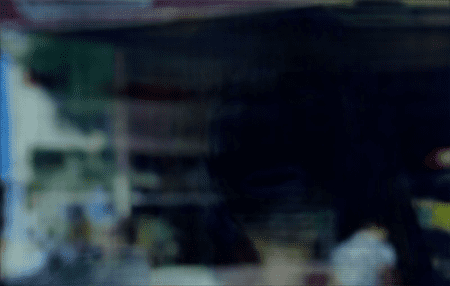}
\vspace{1mm}\\
\includegraphics[width=0.14\linewidth]{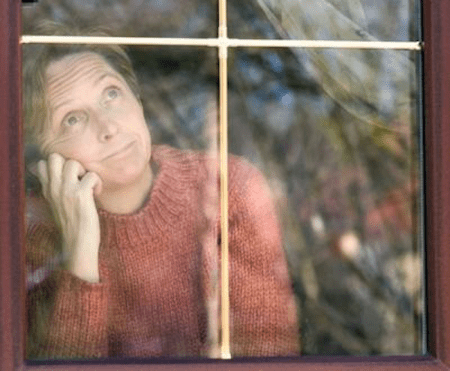}&
\includegraphics[width=0.14\linewidth]{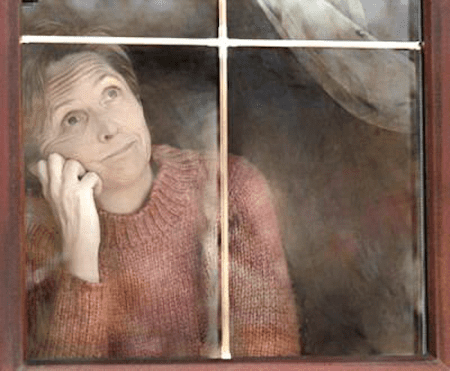}&
\includegraphics[width=0.14\linewidth]{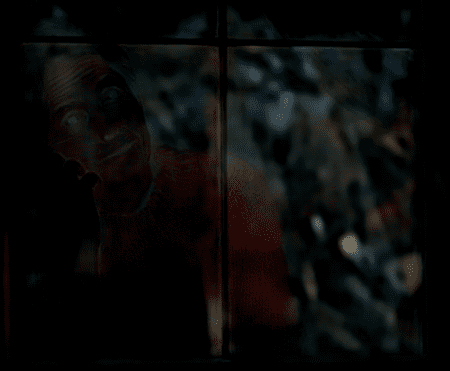}&
\includegraphics[width=0.14\linewidth]{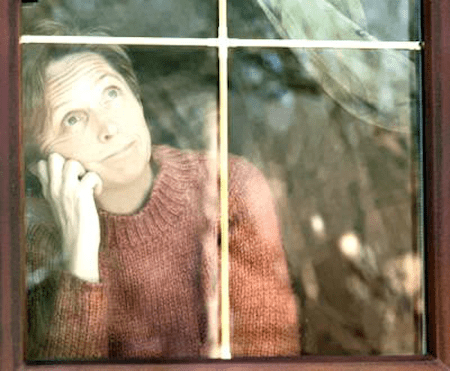}&
\includegraphics[width=0.14\linewidth]{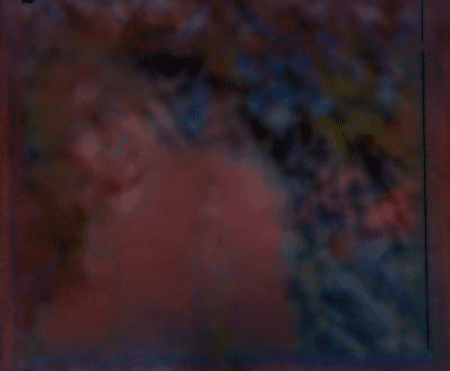}&
\includegraphics[width=0.14\linewidth]{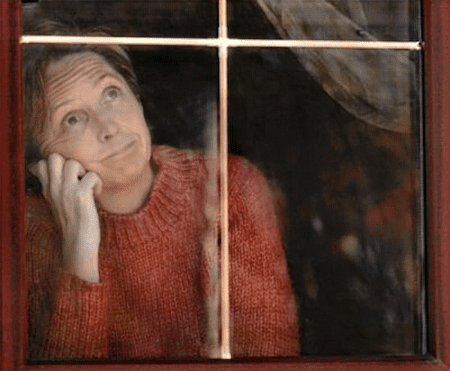}&
\includegraphics[width=0.14\linewidth]{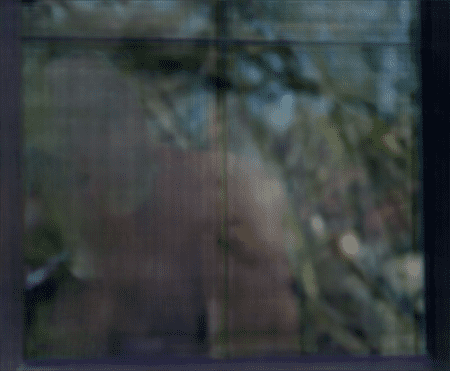}
\vspace{1mm}\\
\includegraphics[width=0.14\linewidth]{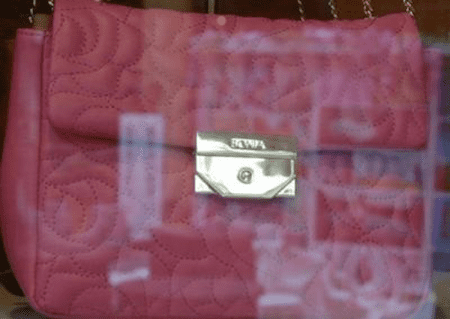}&
\includegraphics[width=0.14\linewidth]{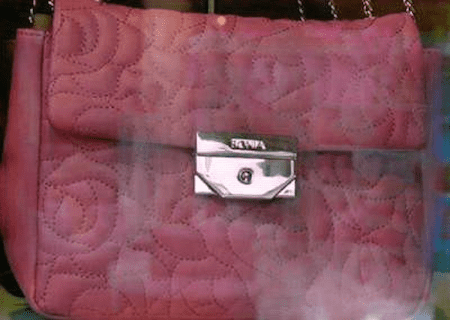}&
\includegraphics[width=0.14\linewidth]{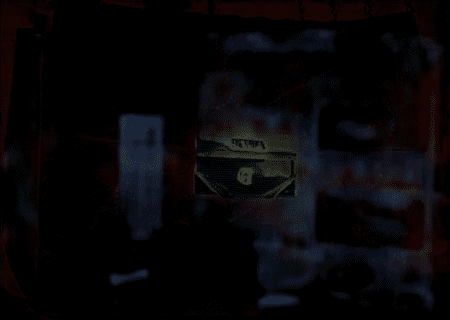}&
\includegraphics[width=0.14\linewidth]{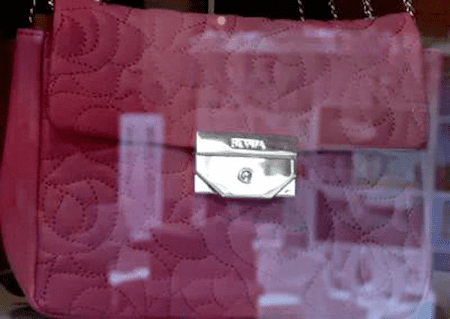}&
\includegraphics[width=0.14\linewidth]{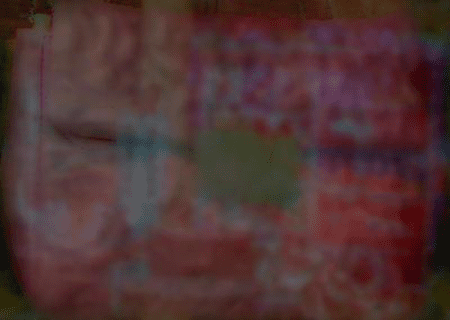}&
\includegraphics[width=0.14\linewidth]{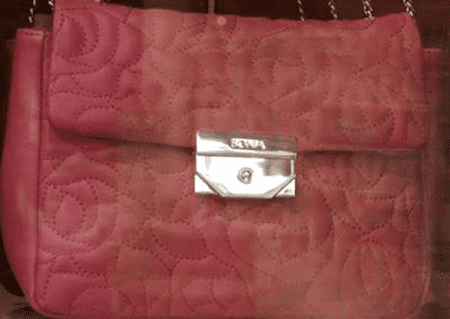}&
\includegraphics[width=0.14\linewidth]{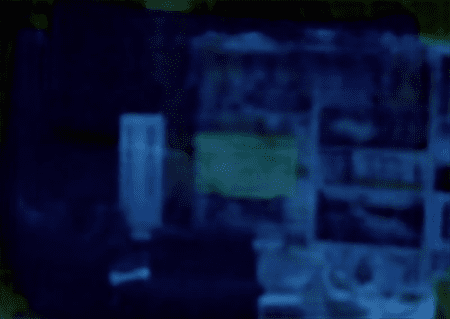}
\vspace{1mm}\\
\small Input & \multicolumn{2}{c}{CEILNet~\cite{Fan2017}} & \multicolumn{2}{c}{Li and Brown~\cite{li2014single}} & \multicolumn{2}{c}{Our results}
\end{tabular}
\caption{Qualitative comparisons among CEILNet~\cite{Fan2017}, Li and Brown~\cite{li2014single} and our method, evaluated on real images in the CEILNET dataset. Note that even though we have no supervision on the reflection layer for real data, but our method predicts cleaner reflection layer as well. Additional results are provided in the supplement.}
\label{fig:qual_real2}
\end{figure*}


\begin{table}[h!]
\centering
\setlength{\tabcolsep}{3mm}
\ra{1.15}
\begin{tabular}{@{}l@{\hspace{6mm}}cc@{\hspace{6mm}}cc@{}}
\toprule
& \multicolumn{2}{c@{\hspace{9mm}}}{Synthetic} & \multicolumn{2}{c}{Real}\\
Method & SSIM & PSNR & SSIM & PSNR \\
\midrule
Ours w/o $L_{\mathrm{feat}}$ & 0.683 & 18.24 & 0.743 & 19.07\\
Ours w/o $L_{\mathrm{adv}}$ & 0.818 & 20.80 & 0.793 & 21.12\\
Ours w/o $L_{\mathrm{excl}}$ & 0.796 & 19.58 & 0.802 & 20.22\\
Ours $L_{\mathrm{adv}}$-only & 0.765 & 18.05 & 0.782 & 19.52\\
\midrule
Ours complete & \textbf{0.853} & \textbf{22.63} & \textbf{0.821} & \textbf{21.30}\\
\bottomrule
\end{tabular}
\caption{Quantitative comparisons on synthetic and real images among multiple ablated models of our method. We remove each of the three losses and evaluate on the re-trained models. 'Ours $L_{\mathrm{adv}}$-only' denotes our method trained with only an adversarial loss. Our complete model shows better performance on both synthetic and real data. We evaluate on synthetic data provided by CEILNet~\cite{Fan2017}, and our real test images described in Section ~\ref{sec:real_data}.}
\label{table:quant_our}
\end{table}
\subsection{Qualitative results}
We present qualitative results of different methods in Figure \ref{fig:qual_real1} and Figure \ref{fig:qual_real2}, evaluated on real-world images from our dataset (with ground truth) and from CEILNet~\cite{Fan2017} (without ground truth), respectively.

\subsection{Controlled experiments}
To analyze how each loss contributes to the final performance of our network, we remove or replace each loss in the combined objective and re-train the network. A visual comparison is shown in Figure \ref{fig:ablation}. When we replace the feature loss $L_{\mathrm{feat}}$ with a $L^1$ loss in color space, the output images tend to be overly-smooth; similar observation is also discussed in ~\cite{Zhao2017,isola2017image}. Without $L_{\mathrm{excl}}$, we notice that visible contents of the reflection layer may appear in the transmission prediction. The adversarial refinement loss $L_{\mathrm{adv}}$ helps recover cleaner and more natural results, as shown in (e).

The quantitative results are shown in Table \ref{table:quant_our}. We also analyze the performance of the model with only an adversarial loss, which is similar to a conditional GAN \cite{isola2017image}.


\begin{figure}[t!]
\centering
\begin{tabular}{@{}c@{\hspace{0.8mm}}c@{\hspace{0.8mm}}c@{}}
\rotatebox{90}{\small \hspace{4mm} Input}&
\includegraphics[width=0.32\linewidth]{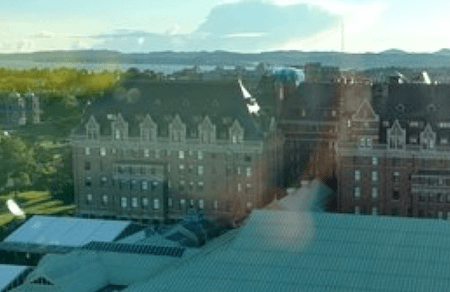}&
\includegraphics[width=0.63\linewidth]{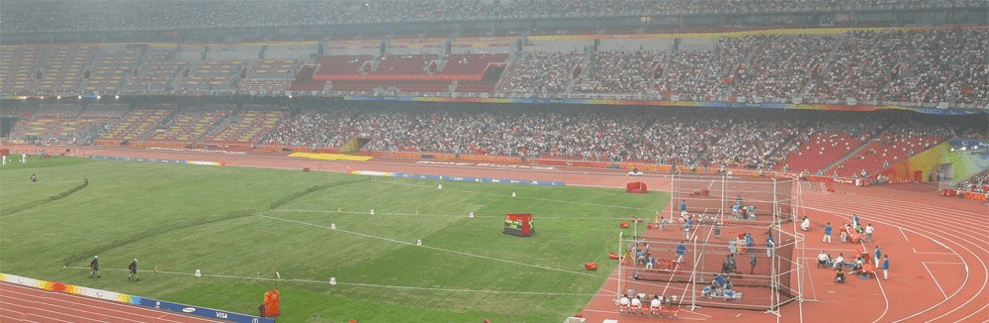}\\
\rotatebox{90}{\small \hspace{5mm} $f_T$}&
\includegraphics[width=0.32\linewidth]{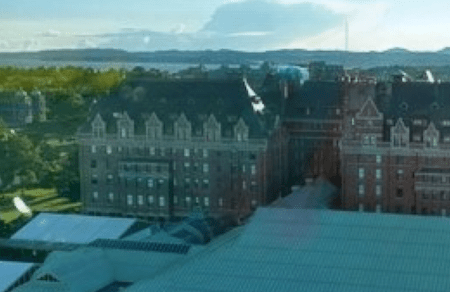}&
\includegraphics[width=0.63\linewidth]{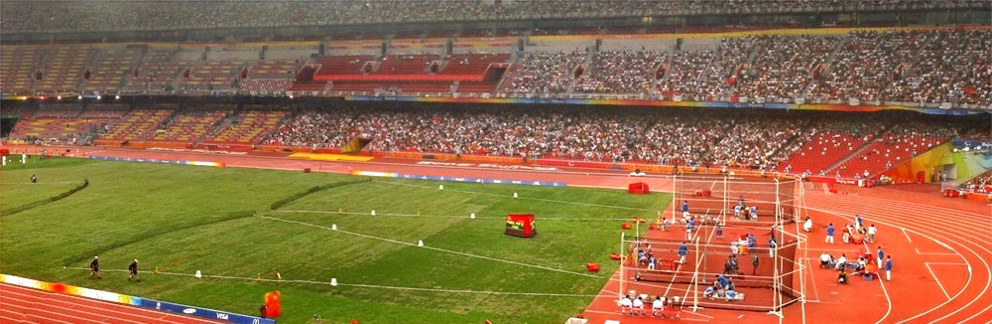}\\
\rotatebox{90}{\small \hspace{5mm} $f_R$}&
\includegraphics[width=0.32\linewidth]{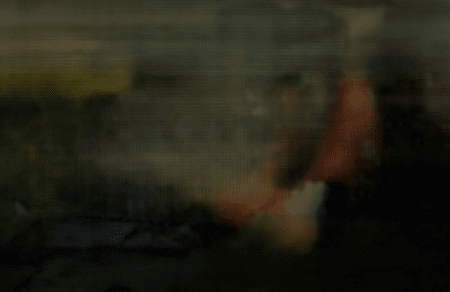}&
\includegraphics[width=0.63\linewidth]{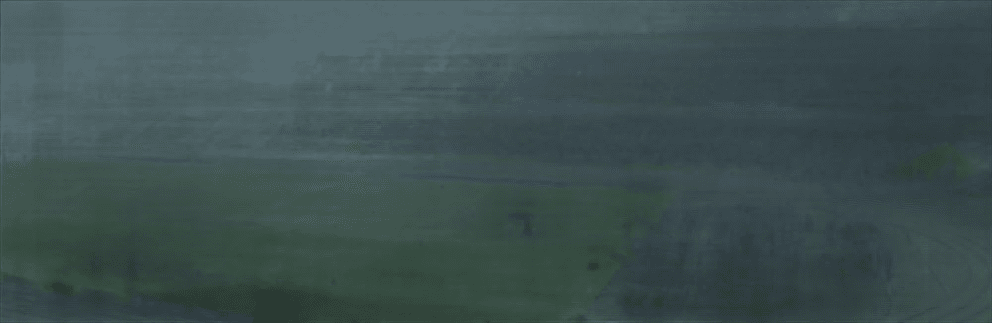}\vspace{1mm}\\
& \small Flare removal & \small Dehazing \vspace{1mm}\\
\end{tabular}
\caption{Extension applications on camera flare removal and image dehazing. For each column, from top to bottom: input, our predicted enhanced layer, our predicted removed layer.}
\label{fig:ext}
\vspace{2mm}
\end{figure}
\section{Extensions}
We demonstrate two additional image enhancement applications, flare removal and dehazing, using our trained model to remove an undesired layer. Note that we directly apply our trained reflection removal model without training or fine-tuning on any flare removal or dehazing dataset. These two tasks can be treated as layer separation problems, similar to reflection separation. For flare removal, we aim to remove the optical artifacts of lens flare, which is caused by light reflection and scattering inside the lens. For dehazing, we target at removing the hazy layer. The hazy images suffer from contrast loss caused by light scattering, reflection and attenuation of particles in the air.  
We show the extension results in Figure \ref{fig:ext}. Our trained model can achieve image enhancement by removing undesirable layers from the input images for flare removal and dehazing. More extension results are provided in the supplement. 

\section{Discussion}
We presented an end-to-end learning approach for single image reflection separation with perceptual losses and a customized exclusion loss. To decompose an image into the transmission and reflection layers, we found it effective to train a network with combined low-level and high-level image features. In order to evaluate different methods on real data, we collected a new dataset of real-world images for reflection removal that contains ground-truth transmission layers. We additionally extend our approach to two other photo enhancement applications to show generality of our approach for layer separation problems.

\begin{figure}
\centering
\begin{tabular}{@{}c@{\hspace{0.8mm}}c@{\hspace{0.8mm}}c@{}}
& Transmission & Reflection\\
\includegraphics[width=0.33\linewidth]{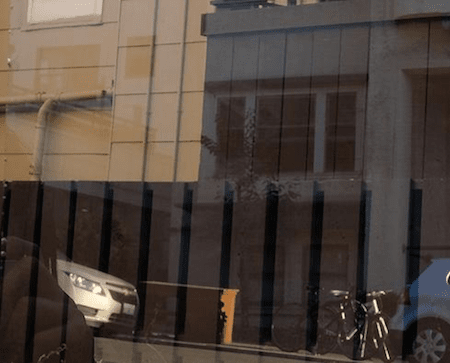}&
\includegraphics[width=0.33\linewidth]{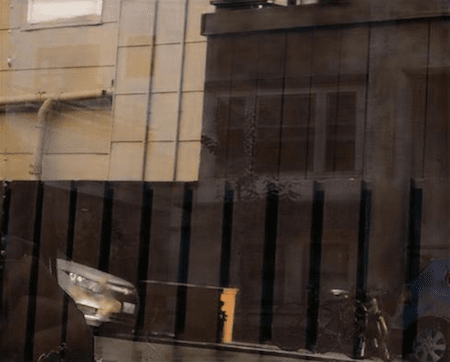}&
\includegraphics[width=0.33\linewidth]{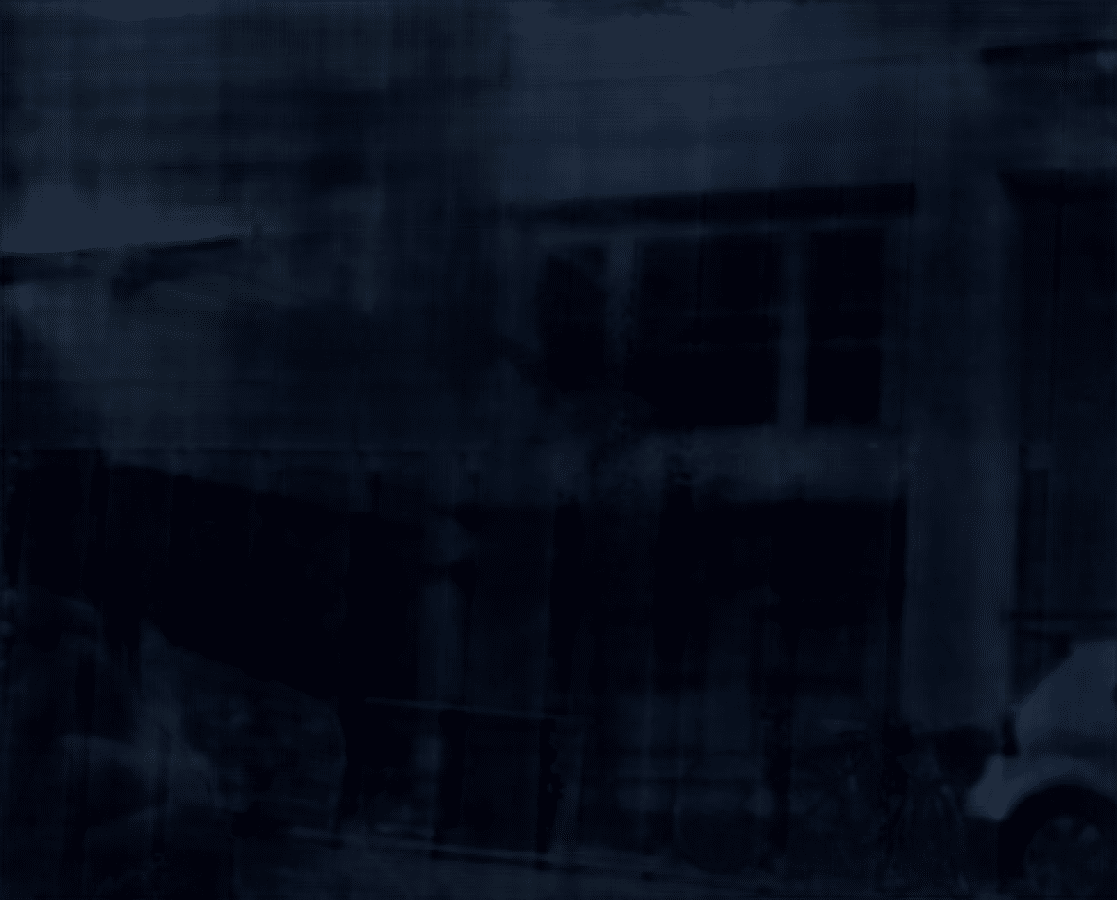}\\
\small Input &\multicolumn{2}{c}{CEILNet~\cite{Fan2017}}\vspace{1mm}\\
\includegraphics[width=0.33\linewidth]{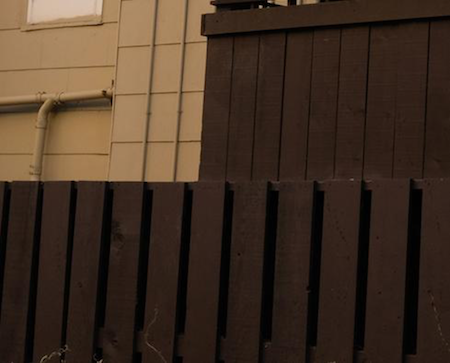}&
\includegraphics[width=0.33\linewidth]{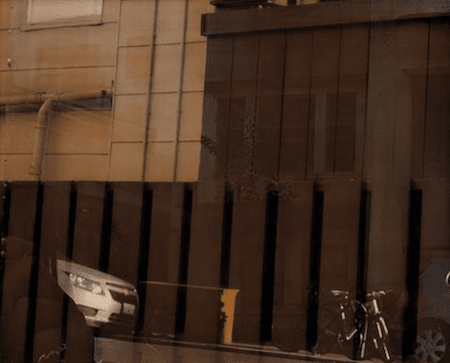}&
\includegraphics[width=0.33\linewidth]{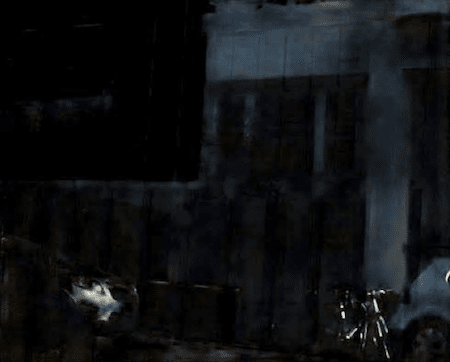}\\
\small Ground-truth T & \multicolumn{2}{c}{Our results} \vspace{1mm}\\
\end{tabular}
\caption{A challenging case with sharp reflection. Our method produces better reflection separation results than CEILNet, but is not able to remove reflection completely.}
\label{fig:limitation}
\vspace{2mm}
\end{figure}

Although our reflection separation model outperforms state-of-the-art approaches on both synthetic and real images, we believe the performance can be further improved in the future. Figure \ref{fig:limitation} illustrates one challenging scenario where the reflection layer is almost as sharp as the transmission layer in a real-world image. We hope our model and dataset will inspire subsequent work on reflection separation and the challenging scenarios. Our dataset and code will be made publicly to facilitate future research.

\section{Acknowledgement}
We thank You Zhang for great help collecting the reflection dataset. We also thank Yichao Zhou and Daniel Seita for constructive writing feedback. This work is supported by UC Berkeley EECS departmental fellowship and hardware donations from NVIDIA.
{\small
\bibliographystyle{ieee}
\bibliography{egbib}
}

\end{document}